\documentclass[runningheads]{llncs}

\usepackage[table,xcdraw]{xcolor}
\usepackage{epsfig}
\usepackage{graphicx}
\usepackage{adjustbox}
\usepackage{amsmath}
\usepackage{amssymb}
\usepackage{booktabs}
\usepackage{multirow}
\usepackage{dsfont}
\usepackage{pifont}
\usepackage[T1]{fontenc}
\usepackage[colorlinks,citecolor=green]{hyperref}
\usepackage{ragged2e}

\begin{document}

\title{\texttt{LayeredDoc}: Domain Adaptive Document Restoration with a Layer Separation Approach}

\titlerunning{LayeredDoc}

\author{Maria Pilligua\inst{1}$^,$\inst{2}\textsuperscript{\textsection}\and
Nil Biescas\inst{1}$^,$\inst{2}\textsuperscript{\textsection} \and
Javier Vazquez-Corral\inst{1}$^,$\inst{2} \and
Josep Llad\'{o}s\inst{1}$^,$\inst{2} \and
Ernest Valveny\inst{1}$^,$\inst{2} \and
Sanket Biswas\inst{1}$^,$\inst{2}}

\authorrunning{M. Pilligua et al.}

\institute{Computer Vision Center \and Computer Science Department \\
Universitat Autònoma de Barcelona, Catalonia, Spain \\
\email{\{mpilligua, jvazquez, josep, ernest, sbiswas\}@cvc.uab.cat} \\
\email{nbiescas@autonoma.cat}}

\maketitle              
\begingroup\renewcommand\thefootnote{\textsection}
\footnotetext{Authors have equally contributed to the Work.}
\endgroup

\begin{abstract}

The rapid evolution of intelligent document processing systems demands robust solutions that adapt to diverse domains without extensive retraining. Traditional methods often falter with variable document types, leading to poor performance. To overcome these limitations, this paper introduces a text-graphic layer separation approach that enhances domain adaptability in document image restoration (DIR) systems. We propose \textit{LayeredDoc}, which utilizes two layers of information: the first targets coarse-grained graphic components, while the second refines machine-printed textual content. This hierarchical DIR framework dynamically adjusts to the characteristics of the input document, facilitating effective domain adaptation. We evaluated our approach both qualitatively and quantitatively using a new real-world dataset, \textit{LayeredDocDB}, developed for this study. Initially trained on a synthetically generated dataset, our model demonstrates strong generalization capabilities for the DIR task, offering a promising solution for handling variability in real-world data. Our code is accessible on this GitHub$^\dagger$

\def\thefootnote{$\dagger$}\footnotetext{\url{https://github.com/mpilligua/LayeredDoc}}

\end{abstract}

\section{Introduction}


Documents, whether they be maps, architectural plans, historical manuscripts, identity documents or administrative paperwork, can be understood as compositions of semantically rich layers. For instance, maps contain layers of geographical information such as mountains, rivers, and roads. Architectural drawings comprise layers that detail structural components, electrical setups, and plumbing systems. Similarly, historical manuscripts include background degradation and foreground components (text and graphic symbols) which are essential for content interpretation as time evolves. The concept of layers also applies to forensic documents (eg. passports, ID cards) because of the multiple objects they have as security control (eg. textures, ultraviolet marks, holograms etc.). Administrative documents are also structured in layers, comprising primary textual content with additional elements like stamps and annotations superimposed. These layers, like transparent overlays, collectively shape the document’s meaning. Their individual recognition characterizes the document within its domain: a particular seal contextualizes the text of the page where it appears. Our hypothesis is that the separation of certain layers of information from the base text is crucial in the adaptation of solving downstream document understanding tasks, mainly Optical Character Recognition (OCR)~\cite{kim2022ocr,souibgui2023text}, Handwritten Text Recognition (HTR)~\cite{kang2022pay,kang2020unsupervised} and Document Layout Analysis (DLA)~\cite{biswas2022docsegtr,biswas2021beyond} over multiple domains.

In the task of document image restoration (DIR), a significant challenge arises from the complex interplay between textual and graphic components within the page, which often include noisy artifacts (eg. ink stains, smears), postage marks (due to stamps, seals), degradation effects (eg. bleedthrough, show-through), background variations (eg. lighting, appearance, shadows), geometric distortion or warps, blurs and watermarks. However, most state-of-the-art (SOTA) approaches have been designed for a specific domain (mostly historical documents) to perform a specific restoration task (mostly document image binarization). This focus on narrow used cases has left a considerable gap in adaptability, especially when dealing with contemporary documents that exhibit a diverse array of artifacts and layout complexities. Consequently, systems that excel in historical document restoration may falter when applied to modern documents, which often feature different types of paper quality, printing techniques, and digital noise. \textit{Our approach aims to bridge this gap by introducing a flexible, domain-adaptive framework that leverages advanced denoising and feature extraction techniques to robustly handle both historical and modern documents across various document processing tasks.} This adaptability is crucial for developing more comprehensive document analysis systems that are capable of performing consistently well across a broad spectrum of document conditions and eras. 

In this work, \textbf{Layer Separation} emerges as a powerful tool to address these challenges by decomposing a document image into distinct layers, each representing different information types such as text, background, and graphics. This segregation not only facilitates enhanced noise reduction and clarity, but could also potentially improve OCR systems by isolating text from disruptive background/foreground elements. In the computer vision literature, layer-wise image decomposition has been employed in tasks such as foreground segmentation ~\cite{yang2022learning}, reflection removal \cite{li2020single}, watermark removal~\cite{liu2021wdnet} and image deraining ~\cite{ren2019progressive}. Inspired by how humans can interact with documents~\cite{liang2005camera} suffering multiple degradations by inferring decompositions between the textual and non-textual components (which include graphical objects, watermarks, and stained artifacts), we propose a simplified two-layered separation of a document image which satisfies the following criteria: (i) the reconstructed layers, which when recombined should yield the original input document. (ii) The reconstructed layers should be independent
of each other (uncorrelated), as they contain their intrinsic properties. This eventually serves as an effective self-supervision strategy to learn robust, generic representations by disentangling the textual and non-textual properties into two separate layers. 

The overall contributions of this work can be divided into 3 folds: (1) A novel text-graphic Layer Separation approach for DIR has been introduced that utilizes a two-layer information processing system. (2) A hierarchical self-supervised DIR framework that allows for effective domain generalization without the need for retraining, making it highly efficient for use in environments where document types and conditions vary frequently. (3) We developed a manually curated real-world evaluation dataset specifically designed for testing the efficacy of unified multi-task DIR frameworks. 

\label{sec:intro}
\section{Related Work}
\subsection{Document Image Restoration}

Many approaches have been proposed to address the enhancement of documents (both handwritten and machine-printed) which suffer several kinds of artefacts/defects such as bleed-through, show-through, faint characters, contrast variations and so on. 
The work from~\cite{calvo2019selectional,kang2021complex} maps images from the degraded domain to the enhanced one using end-to-end CNN-based autoencoders. Other techniques~\cite{jemni2022enhance,souibgui2020gan,souibgui2020conditional} used conditional-Generative Adversarial Network (c-GAN) based approaches to design a generator which produces the enhanced version of the document while the discriminator assesses the quality of binarization. Lately, an end-to-end ViT autoencoder was proposed in~\cite{souibgui2022docentr} to capture high-level global features using self-attention for binarizing degraded documents. Other prominent DIR tasks in the literature mainly include document deblurring~\cite{hradivs2015convolutional}, document dewarping~\cite{das2019dewarpnet,das2022learning}, document deshadowing~\cite{bako2016removing,li2023high}, document relighting or illumination correction~\cite{das2020intrinsic,zhang2023appearance} where restoration models were mainly trained for a specific document enhancement. Efforts to unify all the aforementioned DIR tasks and propose a generalist framework have been lately undertaken by DocRes~\cite{zhang2024docres}. In this work, inspired by ~\cite{zhang2024docres} we propose a more generalist task of Layer Separation for domain adaptive DIR which shows potential to create a unified and flexible DIR framework. 

\subsection{Domain Adaptation for Document Analysis Systems}

Since the data used for pre-training is essentially different from the target domain, domain-adaptive strategies are needed to be considered. In recent years, cross-domain generalization has been studied in DLA task. Li \textit{et al}.~\cite{li2020cross} proposed a benchmark suite that evaluates domain transfer from documents in English language to Chinese and vice-versa.  Recently, Banerjee \textit{et al.}~\cite{banerjee2023swindocsegmenter,banerjee2024semidocseg} proposed a contrastive denoising training objective where a model trained on large collection from scientific domain~\cite{zhong2019publaynet} to other low-resource domains like magazines~\cite{clausner2019icdar2019}. In the handwriting space, Kang \textit{et al.}~\cite{kang2020unsupervised} devised a writer adaptation approach which automatically adjusts a generic handwritten word recognizer, fully trained with synthetic fonts, towards a new incoming writer. Later, they extended the work towards generating synthetic handwritten text lines~\cite{kang2021content}. Synthetic data generation strategies could indeed overcome challenges related to: (i) document source domain: from modern printed to degraded historical samples; (ii) different handwriting styles: from single to multiple writer collections; and (iii) language. Inspired by ~\cite{kang2020unsupervised,kang2021content}, we generate synthetic data which has been used to train our proposed DIR system. This provides a practical and generic approach to deal with unseen and real-world document collections without requiring any retraining or fine-tuning.

\label{sec:related}
\section{Methodology}
\begin{figure}[t!]
    \centering
    \includegraphics[width=\textwidth]{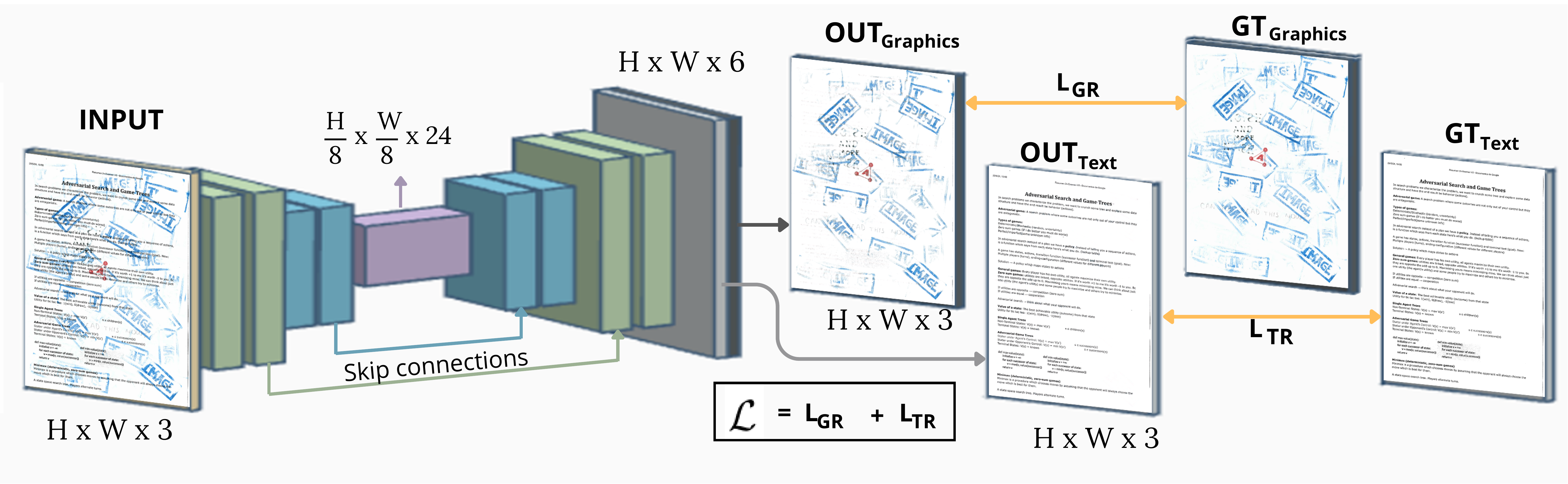}
    \caption{General pipeline of \textbf{LayeredDoc}. We consider the standard architecture of image restoration models and propose to output two image layers instead of the standard single image in those methods. The first layer aims to output the text parts of the document, while the second layer aims to output the overlaid objects.}
    \label{fig:architecure_model}
\end{figure}

In this section, we will explore the proposed adaptation of restoration models for the task of document layer separation, as well as the learning objectives used in the architecture shown in Fig. \ref{fig:architecure_model}.

Our main goal is to adapt current natural image restoration models, mainly based on the U-Net\cite{UNET} architecture, by using a simple yet effective modification in the last upsampling operation, enabling these models to also perform layer separation in documents.

\subsection{Natural Image Restoration}
In natural images restoration, given a degraded image $\mathbf{I} \in \mathbb{R}^{H \times W \times 3}$, current state-of-the-art restoration models such as Restormer \cite{zamir2022restormer} are composed of an encoder $E(x)$ and a decoder $D(x)$. The degraded image $\mathbf{I}$ is passed through the encoder $E(x)$, obtaining latent features $\mathbf{F}_0 \in \mathbb{R}^{\frac{H}{8} \times \frac{W}{8} \times C}$, where $\frac{H}{8} \times \frac{W}{8}$ denotes the spatial dimensions and $C$ is the number of channels. Next, these shallow features $\mathbf{F}_0$ pass through the decoder $D(x)$, which includes several upsampling layers. Finally, a convolution layer is applied to the refined features to generate the output image $\mathbf{R} \in \mathbb{R}^{H \times W \times 3}$. 

\subsection{Layer Separation in Documents} 

In document image processing, the task of layer separation involves isolating distinct components of a document, such as text and overlaid objects, into separate layers. Mathematically, consider an input image \( \mathbf{I} \in \mathbb{R}^{H \times W \times 3} \), where \( H \) and \( W \) denote the height and width of the image, and 3 represents the RGB color channels. The goal is to decompose this image into two separate layers: layer 0 (\( \mathbf{L_0} \)), which may represents the text elements, and layer 1 (\( \mathbf{L_1} \)), which may represent the overlaid objects. 

To do so, we propose to modify natural image restoration architectures to perform text/graphics separation. Our proposed modification happens in the last upsampling layer of these models, as we propose to modify current restoration models to output a  $\mathbf{R'} \in \mathbb{R}^{H \times W \times 6}$ tensor. The first 3 channels of this tensor compose an image that contains the first layer information ($ \mathbf{L_0}$), while the last three channels correspond to the second layer $\mathbf{L_1}$. See Fig. \ref{fig:architecure_model} for a visual explanation.

\subsection{Learning Objectives}


As explained above, we propose that the models output a tensor \( \mathbf{O} \in \mathbb{R}^{H \times W \times 6} \), where the first three channels correspond to the reconstructed layer 0 (\( \mathbf{L_0} \)) and the last three channels correspond to the reconstructed layer 1 (\( \mathbf{L_1} \)). Due to this, we can consider intermediate losses in each of the two layers.

We compute a $L_1$ loss for each of the layers $\mathbf{L_i}$. Mathematically, 
\begin{equation}
    \mathcal{L}_{layer_i}(\mathbf{I_i}, \mathbf{L}_i) = \| {\mathbf{I_i}} - \mathbf{L}_i \|_1,
\end{equation}

where $I_i$ represents the ground truth image for the layer $i$ and ${L}_i$ represents the predicted layer $i$. 

Then, our final loss is just the addition of the losses for both layers:

\begin{equation}
    \mathcal{L}_{final}=\mathcal{L}_{layer_0}+\mathcal{L}_{layer_1}.
\end{equation}

\vspace{3mm}
\paragraph{\textbf{Related Insights:}} 
Our dual loss approach facilitates the model's ability to learn distinct yet complementary tasks. By optimizing for both layer 0 and layer 1 reconstructions, the model gains a richer understanding of the underlying structure within the document images. Thanks to this, and without increasing the model's complexity, our model is able not only to enhance the performance of each individual task but also to leverage the inherent relationship between the two tasks to improve the overall model robustness.




\label{sec:method}
\section{Experiments}

\begin{figure}[t!]
    \centering
    \includegraphics[width=\textwidth]{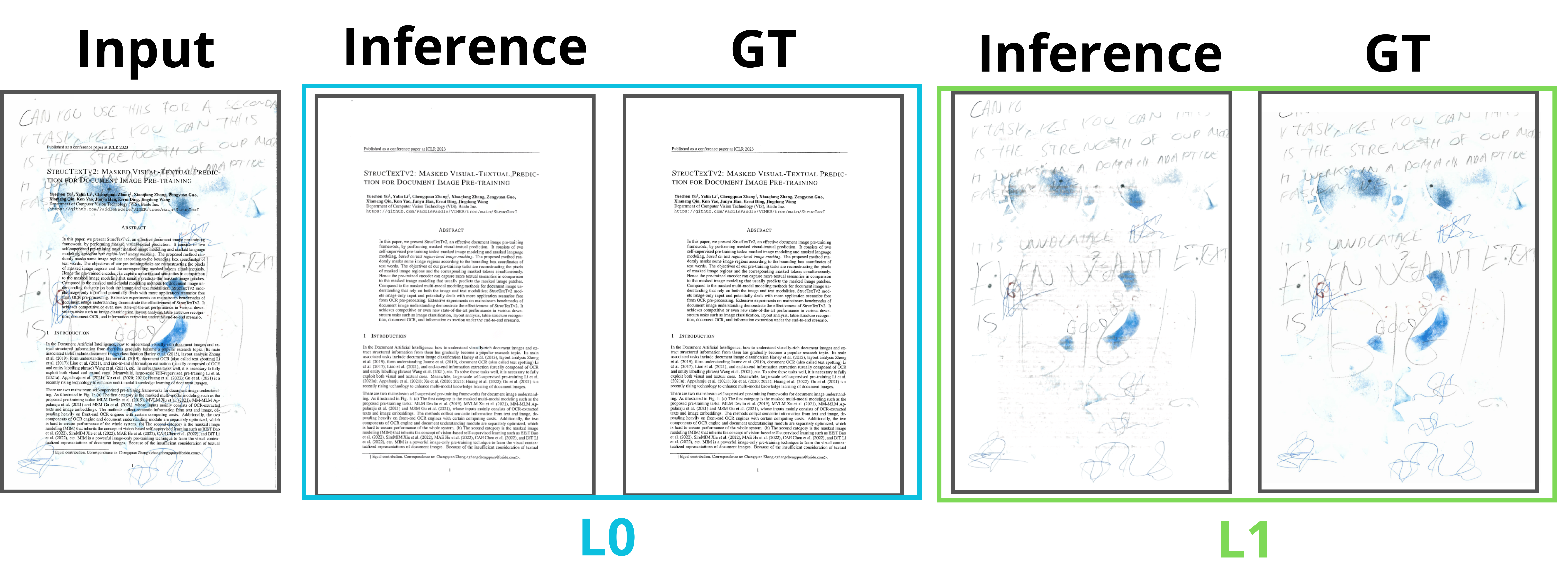}
    \vspace{-7mm}
    \caption{Visual illustration of the layer separation done by our 6-channel LayeredDoc model vs the ground truth}
    \label{fig:6_channels}
\end{figure}

\subsection{Datasets}

\paragraph{\textbf{Training Dataset}} To train our model, we have derived a synthetic dataset. This dataset has the proper ground truth layers to perform supervised learning. We derived ground truth for layer 0 and layer 1, together with the input noisy image. This dataset aims at resembling real  documents containing text with overlayed artifacts like seals and annotations. We selected a subset of Publaynet~\cite{zhong2019publaynet}  with pages that do not contain figures. This set consists of documents with text information only, and therefore these documents build the corresponding Layer 0. Over these images, we have added stamps, signatures, barcodes, QR codes, and passport photos with random rotation, position, and alpha value - which changes the transparency of the image. We have also added random shadows and color changes to simulate that documents were not scanned but captured with a mobile phone.

\begin{figure}[ht]
    \centering
    \includegraphics[width=1\linewidth]{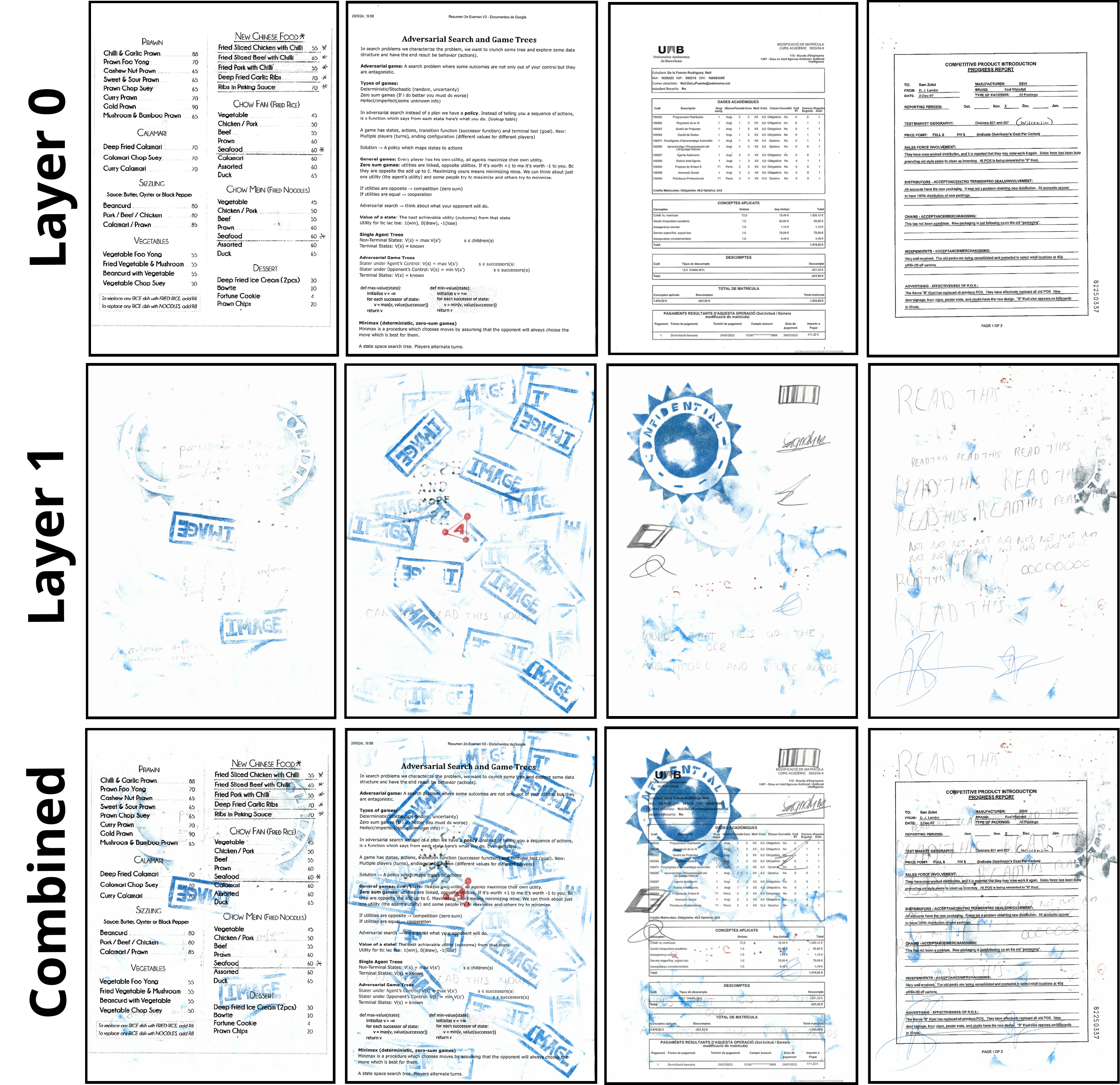}
    \caption{\textbf{Some examples of our manually crafted LayeredDocDB dataset}, illustrating layer 0, layer 1 and the merged noisy image with both layers.}
    \label{fig:our dataset}
\end{figure}

\paragraph{\textbf{LayeredDocDB}} To test our model we crafted a dataset made from real documents. To create this dataset we first randomly added different objects to blank paper sheets. The objects consisted of stamps, signatures and background watermark styled words inspired from~\cite{roy2011document}. These sheets were printed and scanned to include a real-world copy-machine noise to them. Next, we digitally removed the background and, therefore, generated several Layer 1 ($\mathbf{L}_1$) instances. To generate the final synthetic dataset, we overlayed the object images ($\mathbf{L}_1$) to a set of documents coming from a variety of domains, not only those that could be found in Publaynet - scientific papers- to show that our model can adapt to unseen domains.    
In Fig.~\ref{fig:our dataset} we present some examples of our crafted dataset.  

\subsection{Evaluation Metrics}

We evaluated our model using PSNR for color, PSNR for illumination, and SSIM.

\paragraph{\textbf{PSNR-Color:}} Peak Signal-to-Noise Ratio (PSNR) measures the accuracy of color reproduction in the image. Higher PSNR values indicate better color fidelity and overall image quality.

\paragraph{\textbf{PSNR-Illumination:}} This metric assesses the preservation of intensity information in the image. Higher values mean better quality.

\paragraph{\textbf{SSIM (Structural Similarity Index):}} SSIM \cite{SSIM} evaluates perceived image quality by comparing structural information, luminance, and contrast between the original and processed images. It provides a value between -1 and 1, where 1 indicates perfect similarity, reflecting high-quality image preservation.

\subsection{Qualitative Analysis}
Fig.~\ref{fig:6_channels} shows an example of our model considering Restormer \cite{zamir2022restormer} as the baseline restoration architecture. Given a ``noisy'' document image (most-left column), we show the ground truth data and our results for both layers. As we can see,the model is able to correctly decompose the image, presenting the text part of the image on the layer 0 and the overlaid objects (stamps, background text) on the layer 1. This positive separation of text from other objects shall also help in further downstream tasks requiring an OCR, as this is easily confuse by this kind of noise.

\begin{figure}[t!]
    \centering
    \includegraphics[width=1\linewidth]{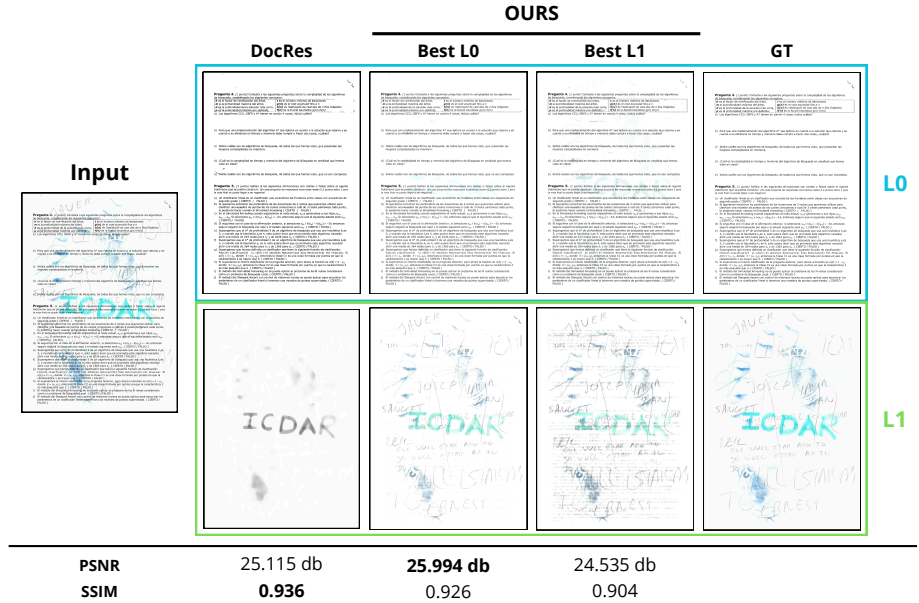}
    \caption{\textbf{Comparison between our LayeredDoc and the DocRes \cite{zhang2024docres} approach.} Our propose framework preserves the color in layer 1 as opposed to DocRes which in comparition puts the objects in gray scale.}
    \label{fig:comp-dores}
\end{figure}

The most similar alternative able to perform this type of layer separation is DocRes \cite{zhang2024docres}. This model also leverages the Restormer image restoration model \cite{zamir2022restormer} by considering a textual prompt. In particular, the DocRes model performs document deshadowing, that can be understand as a similar problem to ours.  Fig.~\ref{fig:comp-dores} shows a visual comparison between the result of DocRes and our results, together with the PSRN and SSIM values obtained for the images in the example. We can see that even though DocRes decently reconstructs the Layer 0 of the document, it fails to obtain a good performance on retrieving Layer 1. This is not the case for the two versions of our model: best $\mathbb{L}_0$ and best $\mathbb{L}_1$ in validation.

\begin{figure}[t!]
    \centering
    \includegraphics[width=1\textwidth]{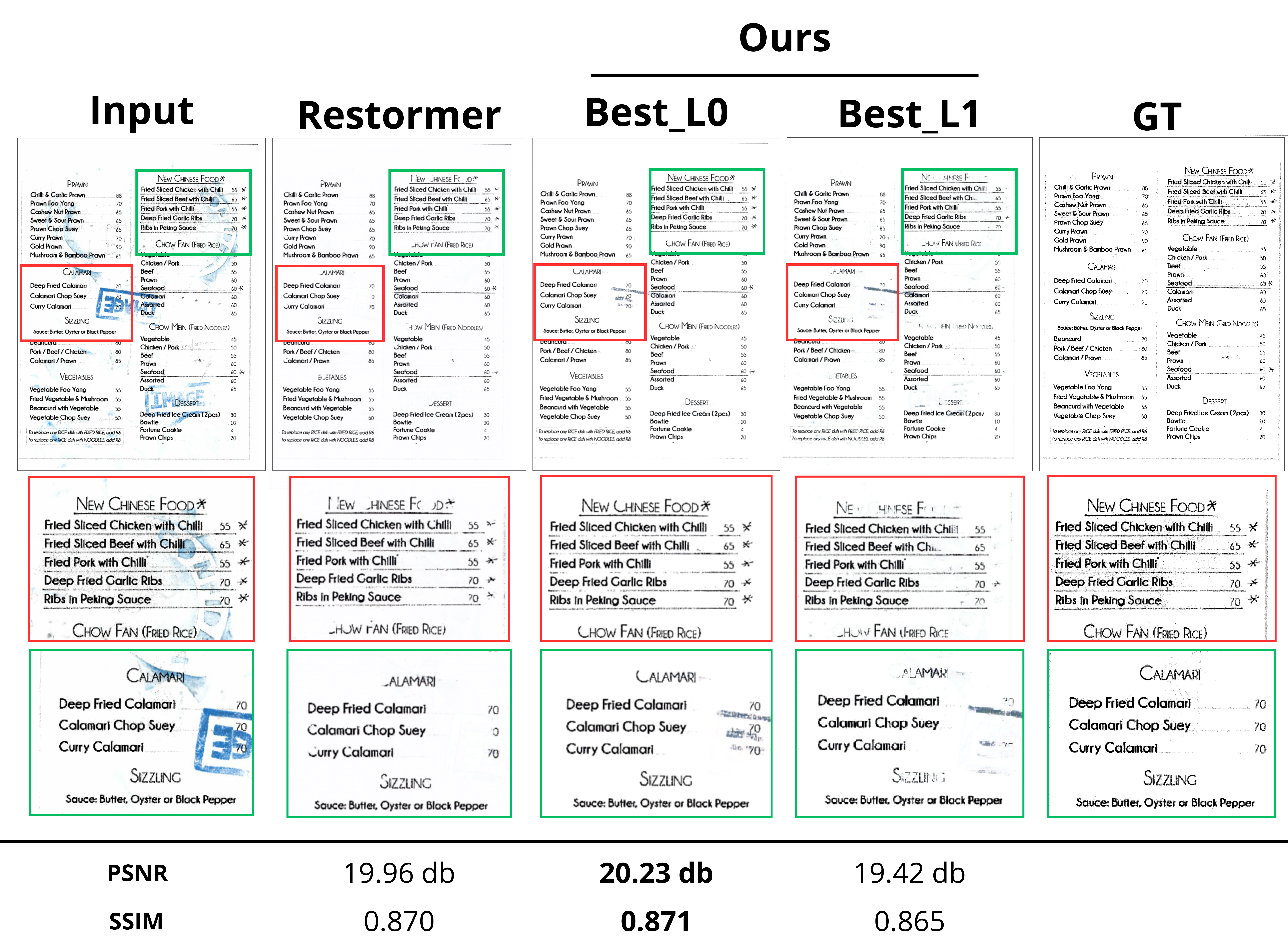}
    \vspace{-3mm}
    \caption{Comparison between the proposed LayeredDoc model and the standard Restormer \cite{zamir2022restormer} approach.}
    \label{fig:enter-label}
\end{figure}

Finally, Fig.~\ref{fig:enter-label} compares the output of our proposed modification versus the output of the standard Restormer model that considers just a 3-channel output. In this figure, we can see how our layer separation model has reconstructed better the document. This is specially noticeable in the titles of the different sections of the menu. While our 6-channel model learns to accurately reconstruct the text, the standard 3-channel model seems to have difficulties and removes parts of the words. As in the previous case, the figure displays PSNR and SSIM results for the images.

\subsection{Quantitative Analysis}
We have also performed a quantitative evaluation comparing the proposed method with DocRes \cite{zhang2024docres} on our synthetically created dataset. The complete summary of results has been depicted in Table \ref{tab:quantitative}. We can see that our model considering the checkpoint with the best $\mathbf{L_0}$ value in validation outperforms both our model with best $\mathbf{L_1}$ and the DocRes model. The results indicate we have a significant \textbf{+2 dB gain} over DocRes in PSNR metrices for both color and illumination. We also have a gain in the SSIM metric showing that LayeredDoc shows a lot of potential over prompt-learning based DocRes approach.

\begin{table}[t!]
\centering
\caption{\textbf{Quantitative Results on our real-world dataset LayeredDocDB}. Final scores computed by averaging over all the images. Results style: \textbf{best}, \underline{second best}.}
\resizebox{0.80\textwidth}{!}{%
\begin{tabular}{|r|r|r|r|r}
\hline
\multicolumn{1}{|l|}{Method} &  \multicolumn{1}{l|}{PSNR(color) ↑} & \multicolumn{1}{l|}{PSNR(ilum) ↑} & \multicolumn{1}{l|}{SSIM ↑} \\ \hline

DocRes~\cite{zhang2024docres}     & 21.2469                            & 22.8686                           & \underline{0.9145} \\        \hline       
\textbf{Ours} (Best\_L0)     & \textbf{23.4026}                            & \textbf{25.0724}                           & \textbf{0.9273}                     \\ \hline
\textbf{Ours} (Best\_L1)    & \underline{21.8596}                            & \underline{23.3913}                           & 0.9034                     \\ \hline
\end{tabular}%
} \label{tab:quantitative}
\vspace{1mm}

\end{table}

\label{sec:experiments}

\newpage
\section{Conclusion and Future Work}
In conclusion, this paper has successfully demonstrated the efficacy of a novel layer separation approach in enhancing the adaptability of document image restoration (DIR) systems across diverse domains. By implementing a dual-layer information processing system, LayeredDoc effectively handles different types of document degradations and complexities without the need for any retraining. The introduction of the LayeredDocDB, a new real-world dataset, further validates our method, showing significant promise in practical applications with its strong generalization capabilities. 

Looking ahead, the future scope of our work could explore the integration of more sophisticated self-supervised frameworks to refine layer separation techniques, potentially increasing accuracy and reducing computational demands. Additionally, expanding the LayeredDocDB to include more varied document types and languages could enhance the robustness and applicability of the DIR system, making it even more effective in global document processing scenarios. Lastly, this work holds a lot of promise to generate more robust and generalist models that could transfer to multiple document understanding tasks in real-world scenario. 
\label{sec:conclusion}

\section*{Acknowledgment}

This work was partially supported by "The European Lighthouse on Safe and Secure AI - ELSA" funded by the European Union’s Horizon Europe programme under grant agreement No 101070617; the Spanish projects PID2021-128178OB-I00 and PID2021-126808OB-I00 funded by  MCIN/AEI/10.13039/501100011033, ERDF "A way of making Europe"; and the Catalan projects 2021-SGR-01499, and 2021-SGR-01559, funded by the Generalitat de Catalunya. S. Biswas is supported by the PhD Scholarship from AGAUR (2023 FI-3-00223), and N. Biescas and M. Pilligua are supported by the CVC Rosa Sensat Student Fellowship. The Computer Vision Center is part of the CERCA Program/Generalitat de Catalunya.
\justifying

\bibliographystyle{splncs04}
\bibliography{main}

\end{document}